\documentclass[journal]{IEEEtran}

\usepackage{graphicx}
\usepackage{amsfonts}
\usepackage{subfigure}
\usepackage[breaklinks=true,letterpaper=true,colorlinks,bookmarks=false]{hyperref}
\usepackage[table]{xcolor}
\usepackage{booktabs}
\usepackage{cite}
\usepackage[cmex10]{amsmath}
\usepackage{gensymb}
\usepackage[small]{caption}
\usepackage{multirow}

\begin{document}
\title{Additional Baseline Metrics for the paper \\ ``Extended YouTube Faces: a Dataset for Heterogeneous Open-Set Face Identification''}
\author{Claudio Ferrari, Stefano Berretti, Alberto Del Bimbo\\
	Media Integration and Communication Center (MICC) \\
	Department of Information Engineering, University of Florence, Italy \\
	{\tt\small \{claudio.ferrari,stefano.berretti,alberto.delbimbo\}@unifi.it }}

\maketitle

\begin{abstract}
In this report, we provide additional and corrected results for the paper ``Extended YouTube Faces: a Dataset for Heterogeneous Open-Set Face Identification''~\cite{ferrari2018extended}.\footnote{The dataset, metadata and protocols are available at \texttt{https://www.micc.unifi.it/resources/datasets/e-ytf/}} After further investigations, we discovered and corrected wrongly labeled images and incorrect identities. This forced us to re-generate the evaluation protocol for the new data; in doing so, we also reproduced and extended the experimental results with other standard metrics and measures used in the literature. The reader can refer to~\cite{ferrari2018extended} for additional details regarding the data collection procedure and recognition pipeline.
\end{abstract}

\appendices

\section{Statistics, Protocols and Metrics}
\subsection{Statistics}
In this appendix, we report the updated statistics of the corrected dataset; $668$ identities out of the original $677$ have been retained, for a total of $37,311$ still images. This is the final number resulting after the filtering procedure. Identities from the original dataset which were less well-known, resulted in uncertain web searches and most of the downloaded images associated to such individuals were wrong. Motivated by the goal of building an open-set protocol, we decided to discard the identities which were considered to contain too many errors. The average number of images per identity is $57$, with a minimum of $1$ image and a maximum of $130$ images. The average video sequence length is instead $180$ frames, with a minimum of $48$ frames and a maximum of $2,157$.
%Table~\ref{tab:datasets} summarizes the characteristics of several face benchmarks in comparison to the E-YTF.

\subsection{Protocols}
The evaluation is to be carried out on 10 splits. For each split, we divide the data into train and test set randomly shuffling the identities, \textit{i.e.}, identities in the train set do not appear in the test set. Following previous works~\cite{klare:2015},  $2/3$ of the identities have been included in the train set, while the remaining $1/3$ constitutes the test set. The train set contains both still images and video frames, while the test set is in turn divided in a probe set and $3$ gallery sets; the latter contain templates of still images (one template per identity), which are defined based on the number of images used to build the template: \textit{(i)}~\textit{Single}: templates of single images, which are selected randomly; \textit{(ii)}~\textit{Half}: templates of half of the total images of the subject, chosen randomly; \textit{(iii)}~\textit{All}: all the available images of the subject are used to build the templates. This aspect is relevant to deepen the impact of differently sized templates in the matching.
The probe set is instead made up of the video sequences.
%each video sequence is used to search the corresponding identity in the gallery.
The search is conducted at video-level, \textit{i.e.}, the decision is taken considering the whole sequence.

The latter setup is used both for the closed-set and the open-set protocols. In the closed-set, the probe identities coincide with the gallery ones; in the open-set, all the identities of the original dataset are used. In this way, some probe subjects do not have a mate in the gallery. Note that in the open-set protocol, the additional probe identities are also disjoint from the training set. In the closed-set, for each split $\sim100$K video frames are used to search into the galleries, which include $\sim12$K, $\sim6$K and $223$ images, respectively, for the \textit{All}, \textit{Half} and \textit{Single} cases. In the open-set, instead, the probe frames are $\sim450$K.
In Table~\ref{tab:datasets-setups} the evaluation setups of some identification benchmark datasets are reported in comparison with the proposed E-YTF.

\begin{table}[!t]
	\centering
	\renewcommand{\arraystretch}{1.1}
	\caption{Evaluation setups of identification benchmark datasets}\label{tab:datasets-setups}
	\scalebox{0.87}{
		\begin{tabular}{lrrcrr}
			\hline
			Dataset & \#ID  & \#train-ID & evaluation & \#gallery-ID & \#probe-ID \\
			\hline  		
			MegaFace~\cite{kemelmacher-shlizerman:2016} & 690K  & -- & 2 probe sets & 690K & 530 / 975 \\		
			IJB-A~\cite{klare:2015} & 500  & 333 & 10 splits & 167 & 167 \\
			IJB-B~\cite{whitelam:2017} & 1,845 & -- & 2 gallery sets & 931 / 914 & 1.845 \\  		
			\hline
			\bf{E-YTF (closed)} & \bf{668}  & \bf{450} & \bf{10 splits} & \bf{223} & \bf{223} \\
			\hline
			\bf{E-YTF (open)} & \bf{1595}  & \bf{450} & \bf{10 splits} & \bf{223} & \bf{1149}
			\\ \hline
		\end{tabular}}
\end{table}
	
\subsection{Metrics}	
In addition to the metrics reported in~\cite{ferrari2018extended}, \textit{i.e.}, CMC and ROC curves for the closed-set, IET curve for the open-set, in this report we provide additional measures commonly used in literature. More specifically:

\begin{figure*}[!t]
	%\centering
	\includegraphics[width=0.245\linewidth]{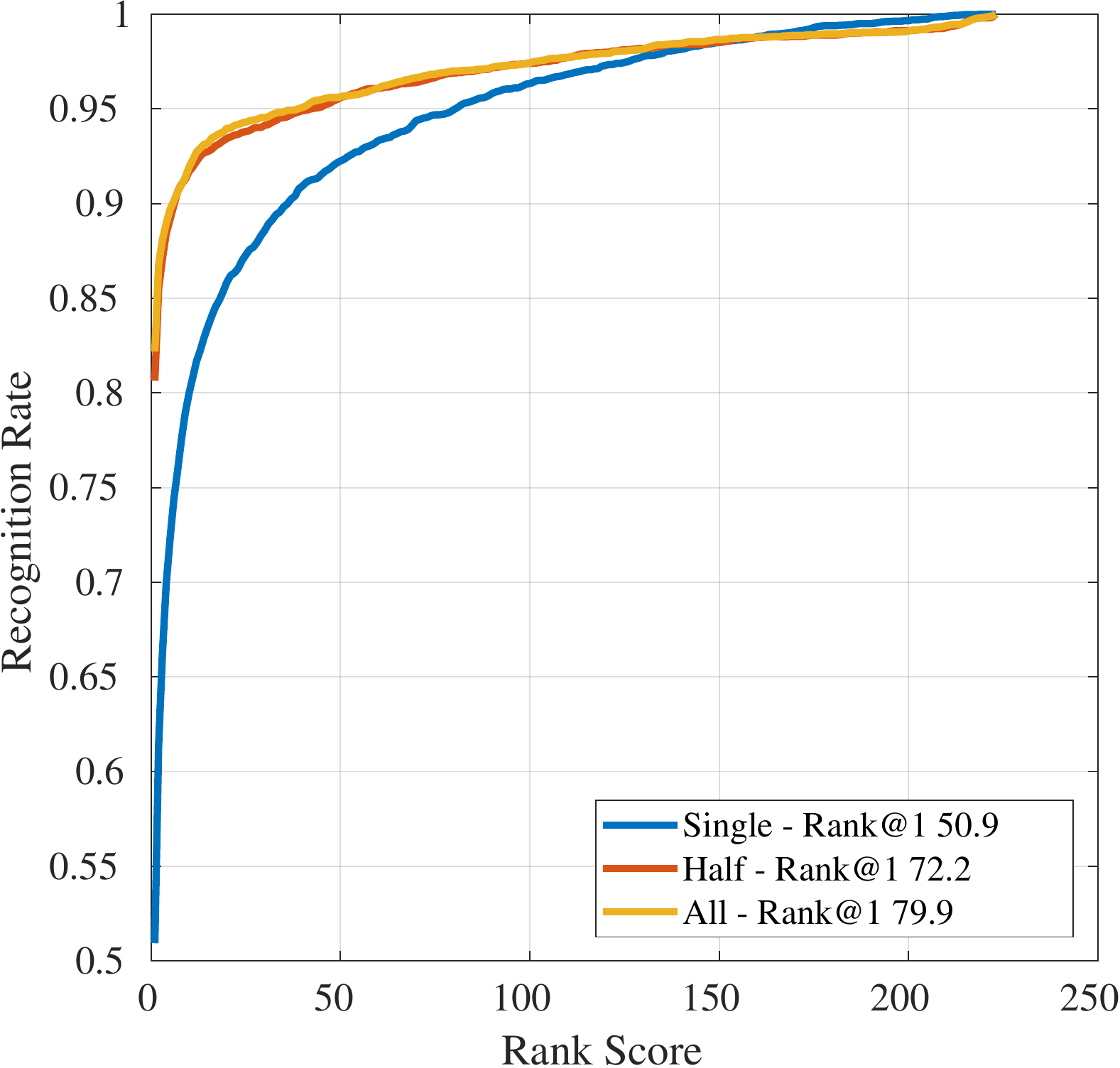}			
	\includegraphics[width=0.245\linewidth]{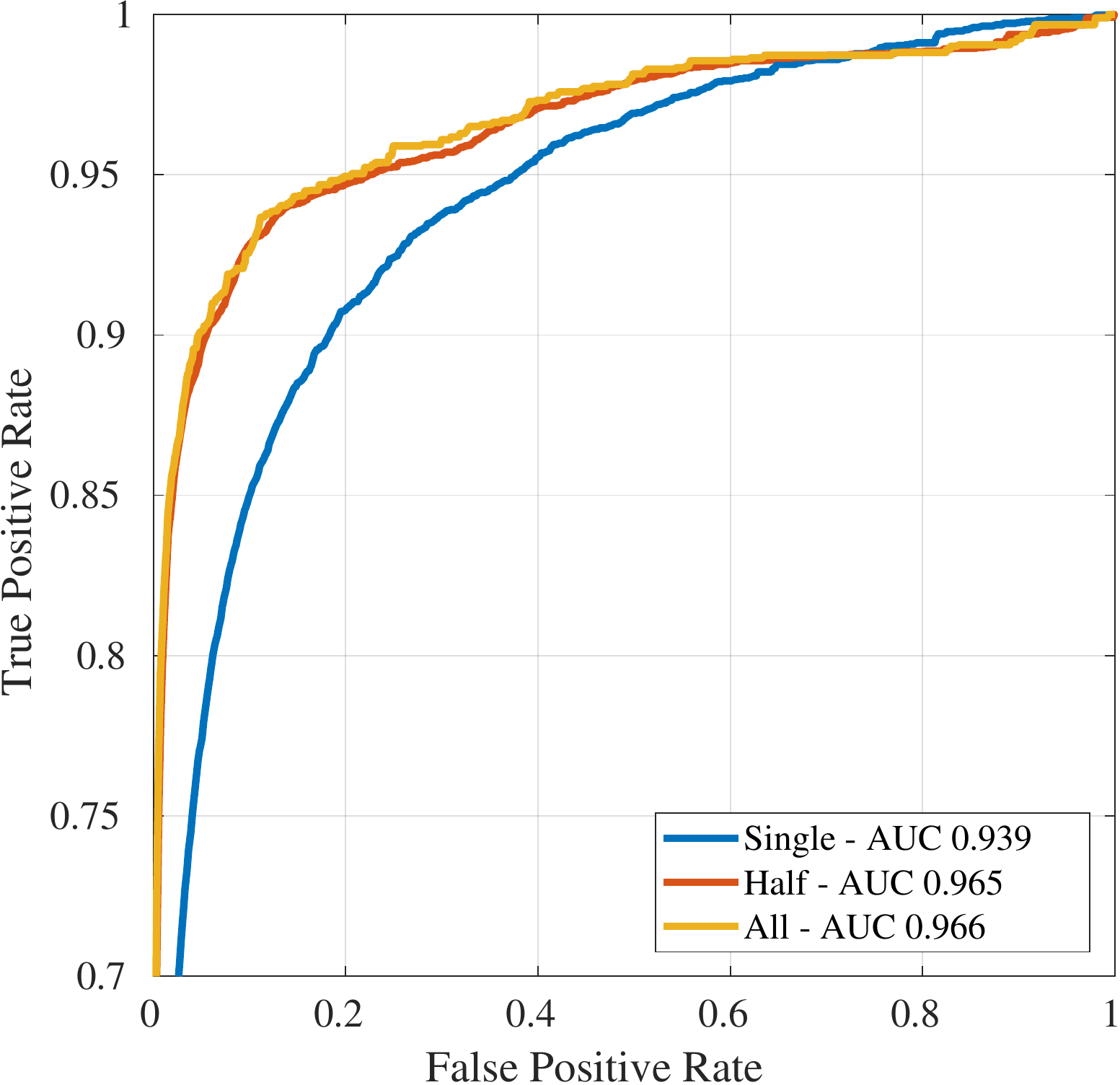}
	\includegraphics[width=0.245\linewidth]{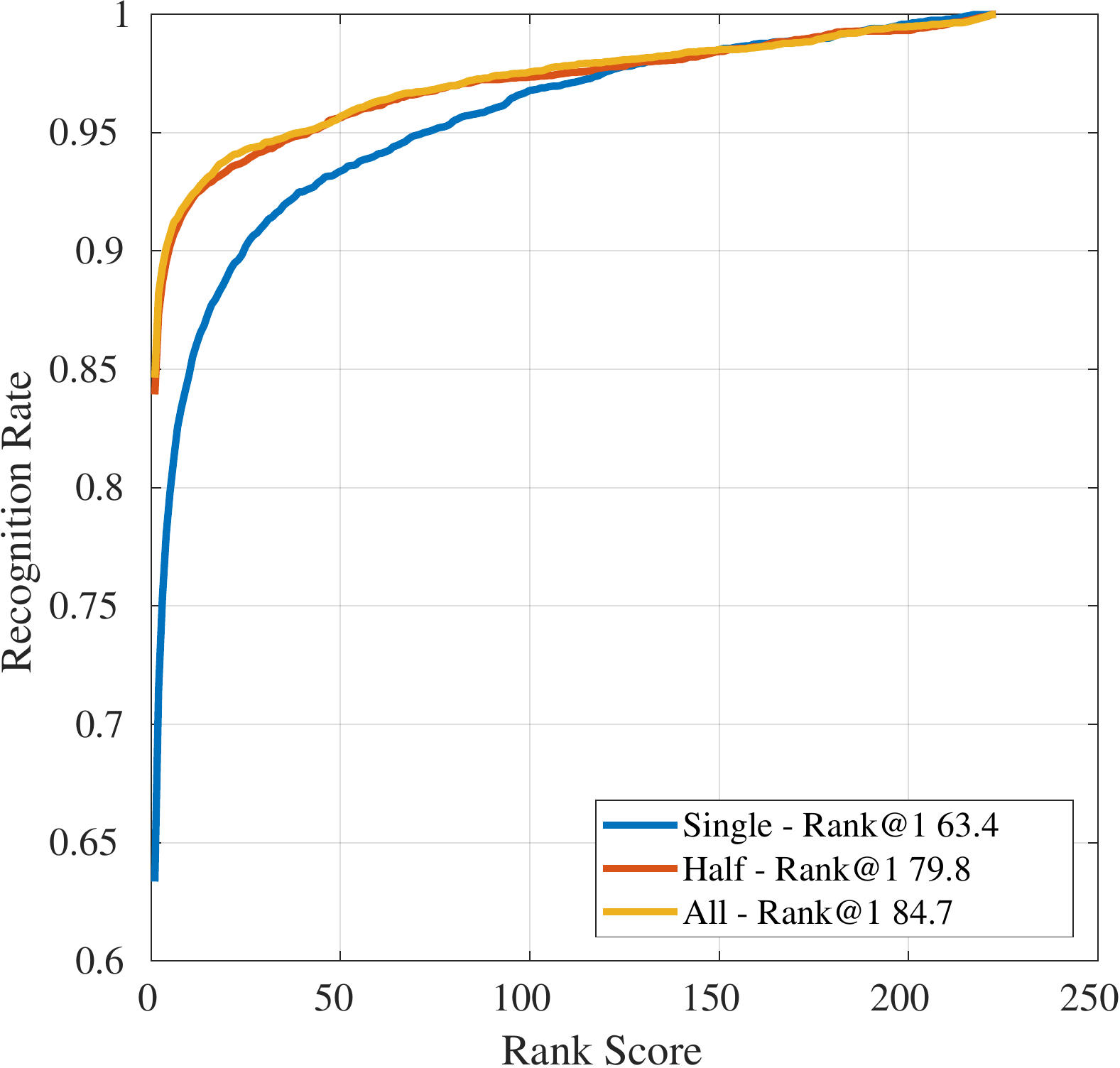}	
	\includegraphics[width=0.245\linewidth]{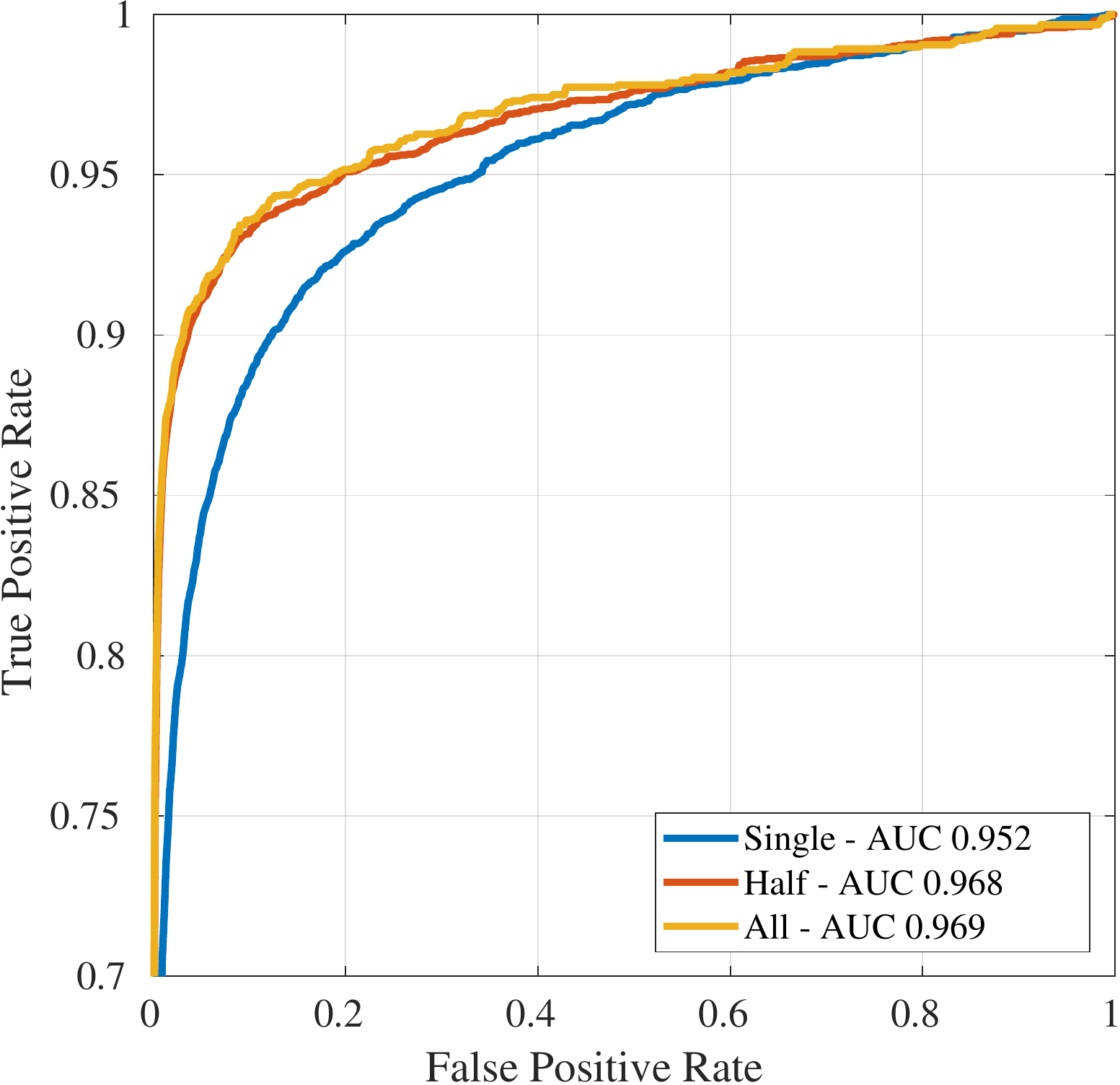}	
	\begin{minipage}{1\linewidth}
		\hspace{60pt} (a)
		\hspace{111pt} (b)
		\hspace{112pt} (c)
		\hspace{112pt} (d)						
	\end{minipage}	
	\caption{Average CMC and ROC curves across 10 splits for AlexNet (a-b) and VggFace (c-d), and for the different templates size.}
	\label{fig:cmc-roc-closed}
\end{figure*}

\begin{table*}[!ht]
	\centering
	\renewcommand{\arraystretch}{1.1}
	\caption{Rank@1, Rank@10 and TAR@FAR rates in function of the template size for the Closed-set scenario}\label{tab:closed-set}
	\scalebox{0.99}{
		\begin{tabular}{c|c|cc|cccc}
			
			Template & Net & Rank@1 & Rank@10 & TAR@$10^{-2}$FAR & TAR@$10^{-3}$FAR & TAR@$10^{-4}$FAR & TAR@$10^{-5}$FAR \\
			\hline
			
			\multirow{ 2}{*}{Single} & VggFace & $0.634 \pm 0.018$ & $0.847 \pm 0.014$ & $0.714 \pm 0.008$ & $0.489 \pm 0.010$ & $0.257 \pm 0.012$ & $0.124 \pm 0.018$ \\
			& AlexNet & $0.509 \pm 0.033$ & $0.799 \pm 0.016$ & $0.589 \pm 0.007$ & $0.349 \pm 0.012$ & $0.175 \pm 0.012$ & $0.080 \pm 0.021$ \\
			\hline  		
			\multirow{ 2}{*}{Half} & VggFace & $0.839 \pm 0.020$ & $0.920 \pm 0.016$ & $0.857 \pm 0.012$ & $0.707 \pm 0.009$ & $0.537 \pm 0.012$ & $0.305 \pm 0.056$\\
			& AlexNet & $0.807 \pm 0.015$ & $0.917 \pm 0.015$ & $0.798 \pm 0.011$ & $0.573 \pm 0.015$ & $0.355 \pm 0.012$ & $0.163 \pm 0.015$\\
			\hline  		
			\multirow{ 2}{*}{All} & VggFace & $\mathbf{0.846 \pm 0.017}$ & $\mathbf{0.922 \pm 0.013}$ & $\mathbf{0.859 \pm 0.010}$ & $\mathbf{0.727 \pm 0.010}$ & $\mathbf{0.547 \pm 0.010}$ & $\mathbf{0.311 \pm 0.051}$\\
			& AlexNet &  $0.822 \pm 0.014$ & $0.919 \pm 0.014$ & $0.809 \pm 0.010$ & $0.600 \pm 0.017$ & $0.380 \pm 0.006$ &$0.188 \pm 0.020$\\		
			\hline
		\end{tabular}}
\end{table*}

\subsubsection{Closed-set} we report Rank@1, Rank@10 recognition rates and True Acceptance Rate (TAR) for different False Acceptance Rates (FAR).

\subsubsection{Open-set} we report True Positive Identification Rate (TPIR) for different False Positive Identification Rate (FPIR). This measure is widely employed to assess the accuracy of the model with respect to its robustness in rejecting impostors. FPIR is computed over K searches, each involving imagery from a person who is known to not to be present in the enrolled gallery, and is defined as the proportion of searches with any candidates at or above/below a score/distance threshold $T$. In order to compare different approaches, performance are computed at fixed FPIR values on the impostor candidates. The threshold $T$ is selected such that the FPIR is below the fixed rates, and is then used on the test set to filter the impostors and rank the probe queries. The TPIR is expressed in terms of Rank@1 recognition;

\subsection{Pipeline}	
Both the still images and the video sequences come along with bounding box annotations, thus the detection step was skipped. Following the guidelines in~\cite{ferrari2018investigating}, the provided bounding boxes were enlarged so as to include the whole head and the alignment step was bypassed too. The face crops and their horizontally flipped version were then fed to two different pre-trained CNN architectures to extract feature descriptors; the final descriptor is obtained as the average of the two. We employed the publicly available VggFace model~\cite{parkhi:2015} and the AlexNet architecture~\cite{krizhevsky2012imagenet}, trained as in~\cite{ferrari2018investigating}.
For each video sequence in the probe set, we computed the average descriptor from all the frames. The motivation for this is two-fold: first, the YTF video sequences are rather short and thus the variability in the appearance is supposed to be limited; in this sense, it also helps in attenuating the effect of outliers. Secondly, it allows a much faster matching procedure.
	
Finally, we employed the cosine distance to match probe and gallery. Being the gallery composed of templates, one needs to derive a single scalar value from all the distances computed between the video sequences and each image in the templates. All the following reported results are presented in terms of the ``\textit{min+mean}`` distance metric as described in~\cite{ferrari2018investigating}.

\section{Closed-set Recognition Results}
In Table~\ref{tab:closed-set} and Figure~\ref{fig:cmc-roc-closed}, we report results as a function of the gallery template sizes for the AlexNet and VggFace architectures. Results evidence that the gallery template size has a noticeable impact on the performance; the additional images in the ``half`` and `all'' cases provide significant information, while the system still struggles in handling a single gallery image. Matching heterogeneous data adds further difficulty to the protocol, with the best reported result of $84,6\%$ at Rank@1 recognition rate.

\section{Open-set Recognition Results}
\begin{figure}[!t]
	%\centering
	\includegraphics[width=0.49\linewidth]{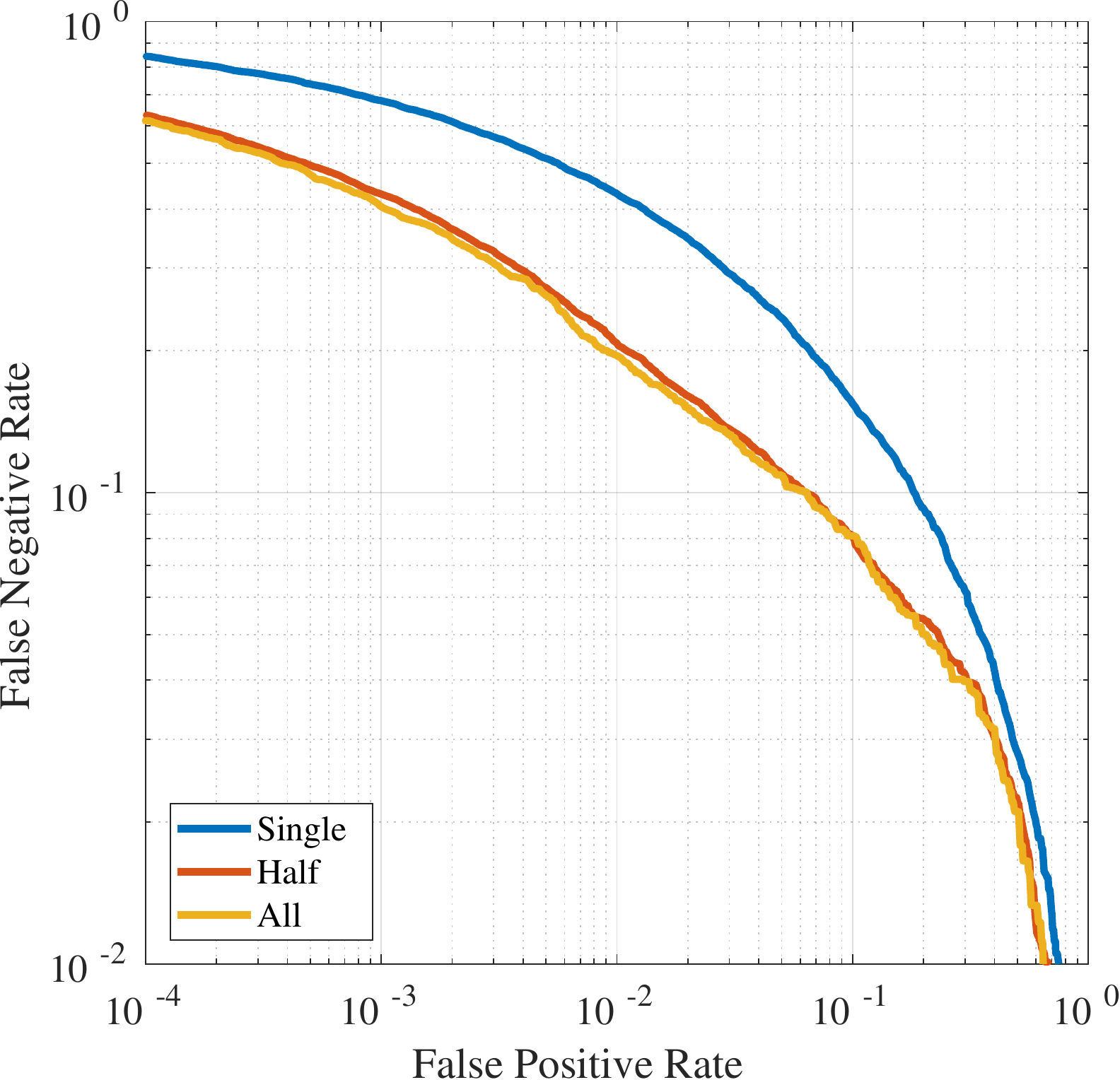}	
	\includegraphics[width=0.49\linewidth]{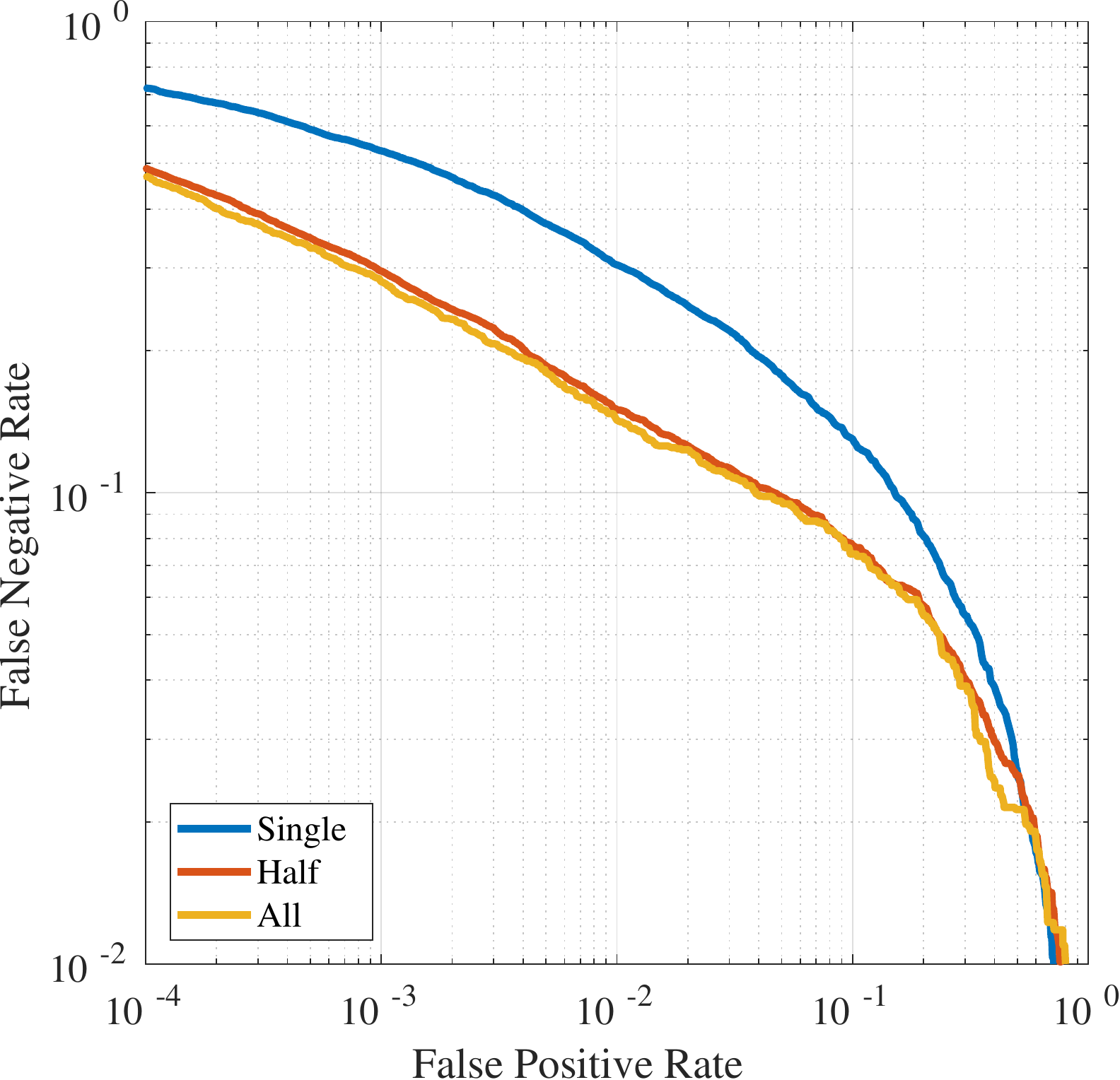}			
	\begin{minipage}{1\linewidth}
		\hspace{57pt} (a)
		\hspace{111pt} (b)					
	\end{minipage}	
	\caption{Average IET performance across 10 splits for AlexNet (a) and VggFace (b), and for the different templates size.}
	\label{fig:iet-open}
\end{figure}

Table~\ref{tab:open-set} and Figure~\ref{fig:iet-open} report results for the open-set protocol, in terms of TPIR at different FPIR rates and IET performance curve. Outcomes for this protocol evidence the difficulty of handling such a large number of impostor identities ($926$), posing further challenges in open-set face recognition.

\begin{table}[!ht]
	\centering
	\renewcommand{\arraystretch}{1.1}
	\caption{Results for the Open-set scenario}\label{tab:open-set}
	\scalebox{0.95}{
		\begin{tabular}{c|c|ccc}
			
			Template & Net & FPIR $10^{-1}$ & FPIR $10^{-2}$ & FPIR $10^{-3}$\\
			\hline
			
			\multirow{ 2}{*}{Single} & VggFace & $0.413 \pm 0.024$ & $0.203 \pm 0.027$ & $0.111 \pm 0.035$ \\
			& AlexNet & $0.269 \pm 0.021$ & $0.123 \pm 0.018$ & $0.064 \pm 0.019$ \\
			\hline  		
			\multirow{ 2}{*}{Half} & VggFace & $0.664 \pm 0.018$ & $0.431 \pm 0.033$ & $0.248 \pm 0.048$ \\
			& AlexNet & $0.535 \pm 0.022$ & $0.300 \pm 0.033$ & $0.146 \pm 0.033$ \\
			\hline  		
			\multirow{ 2}{*}{All} & VggFace & $0.682 \pm 0.020$ & $0.448 \pm 0.020$ & $0.269 \pm 0.033$\\
			& AlexNet & $0.561 \pm 0.014$ & $0.317 \pm 0.030$ & $0.172 \pm 0.017$ \\					
		\end{tabular}}
	\end{table}

\section{Conclusions}
In this short report, we corrected and integrated baseline results for the proposed ``Extended YouTube Faces'' dataset~\cite{ferrari2018extended}, adding standard evaluation metrics to ease the comparison with other datasets and approaches.

\section{Acknowledgments}
The Titan Xp used for this research was donated by the NVIDIA Corporation.

\bibliographystyle{plain}

\begin{thebibliography}{1}
	
	\bibitem{ferrari2018extended}
	Claudio Ferrari, Stefano Berretti, and Alberrto Del~Bimbo.
	\newblock Extended youtube faces: a dataset for heterogeneous open-set face
	identification.
	\newblock In {\em 2018 24th International Conference on Pattern Recognition
		(ICPR)}, pages 3408--3413. IEEE, 2018.
	
	\bibitem{ferrari2018investigating}
	Claudio Ferrari, Giuseppe Lisanti, Stefano Berretti, and Alberto Del~Bimbo.
	\newblock Investigating nuisances in dcnn-based face recognition.
	\newblock {\em IEEE Transactions on Image Processing}, 27(11):5638--5651, 2018.
	
	\bibitem{kemelmacher-shlizerman:2016}
	I.~Kemelmacher-Shlizerman, S.~M. Seitz, D.~Miller, and E.~Brossard.
	\newblock The megaface benchmark: 1 million faces for recognition at scale.
	\newblock In {\em IEEE Conf. on Computer Vision and Pattern Recognition
		(CVPR)}, pages 4873--4882, June 2016.
	
	\bibitem{klare:2015}
	B.~F. Klare, B.~Klein, E.~Taborsky, A.~Blanton, J.~Cheney, K.~Allen,
	P.~Grother, A.~Mah, M.~Burge, and A.~K. Jain.
	\newblock Pushing the frontiers of unconstrained face detection and
	recognition: Iarpa janus benchmark a.
	\newblock In {\em IEEE Conf. on Computer Vision and Pattern Recognition
		(CVPR)}, pages 1931--1939, June 2015.
	
	\bibitem{krizhevsky2012imagenet}
	Alex Krizhevsky, Ilya Sutskever, and Geoffrey~E Hinton.
	\newblock {ImageNet} classification with deep convolutional neural networks.
	\newblock In {\em Int. Conf. on Advances in Neural Information Processing
		Systems (NIPS)}, pages 1097--1105, 2012.
	
	\bibitem{parkhi:2015}
	Omkar~M. Parkhi, Andrea Vedaldi, and Andrew Zisserman.
	\newblock Deep face recognition.
	\newblock In {\em British Machine Vision Conf. (BMVC)}, volume~1, page~6, 2015.
	
	\bibitem{whitelam:2017}
	C.~Whitelam, E.~Taborsky, A.~Blanton, B.~Maze, J.~Adams, T.~Miller, N.~Kalka,
	A.~K. Jain, J.~A. Duncan, K.~Allen, J.~Cheney, and P.~Grother.
	\newblock Iarpa janus benchmark-b face dataset.
	\newblock In {\em IEEE Conf. on Computer Vision and Pattern Recognition
		Workshops}, pages 592--600, 2017.
	
\end{thebibliography}

\end{document}